\documentclass[pmlr]{jmlr}

\usepackage{longtable}

\usepackage{booktabs}
\usepackage{cleveref}
\usepackage{xcolor}
\usepackage{comment}
\usepackage{wrapfig}

\usepackage[load-configurations=version-1]{siunitx} 

\theorembodyfont{\upshape}
\theoremheaderfont{\scshape}
\theorempostheader{:}
\theoremsep{\newline}

\newcommand{\sm}[1]{{\textbf {\color{orange} SM: #1}}}

\jmlrvolume{}
\jmlryear{}
\jmlrworkshop{Under review}

\title[MineRL Diamond 2021 Competition]{MineRL Diamond 2021 Competition: Overview, Results, and Lessons Learned}

\author{
 \Name{Anssi Kanervisto}\thanks{Equal contribution} \Email{anssk@uef.fi} \\
 \addr University of Eastern Finland
 \AND
 \Name{Stephanie Milani}\footnotemark[1] \Email{smilani@cs.cmu.edu} \\
 \addr Carnegie Mellon University
 \AND
 \Name{Karolis Ramanauskas} \Email{karolis.ram@gmail.com} \\
 \addr Independent
 \AND
 \Name{Nicholay Topin} \Email{ntopin@cs.cmu.edu} \\
 \addr Carnegie Mellon University
 \AND \\
 \Name{Zichuan Lin} \Email{zichuanlin@tencent.com} \\
 \Name{Junyou Li} \Email{junyouli@tencent.com} \\
 \Name{Jianing Shi} \Email{jianingshi@tencent.com} \\
 \Name{Deheng Ye} \Email{dericye@tencent.com} \\
 \Name{Qiang Fu} \Email{leonfu@tencent.com} \\
 \Name{Wei Yang} \Email{willyang@tencent.com} \\
 \addr Tencent AI Lab \\
 \AND
 \Name{Weijun Hong} \Email{hongweijun@corp.netease.com} \\
 \Name{Zhongyue Huang} \Email{huangzhongyue@corp.netease.com} \\
 \Name{Haicheng Chen} \Email{chenhaicheng@corp.netease.com} \\
 \Name{Guangjun Zeng} \Email{gzzengguangjun@corp.netease.com} \\
 \Name{Yue Lin} \Email{gzlinyue@corp.netease.com} \\
 \addr NetEase Games AI Lab \\
 \AND
 \Name{Vincent Micheli} \Email{vincent.micheli@unige.ch} \\
 \Name{Eloi Alonso} \Email{eloi.alonso@unige.ch} \\
 \Name{Fran\c{c}ois Fleuret} \Email{francois.fleuret@unige.ch} \\
 \addr University of Geneva \\
 \AND
 \Name{Alexander Nikulin} \Email{hsehowuhh@gmail.com} \\
 \Name{Yury Belousov} \Email{yury.belousov@unige.ch} \\
 \Name{Oleg Svidchenko} \Email{oleg.svidchenko@jetbrains.com} \\
 \Name{Aleksei Shpilman} \Email{aleksei@shpilman.com} \\
 \addr HSE University, JetBrains Research
}

\editor{}

\newcommand{\easy}{%
\texttt{intro}%
}
\newcommand{\hard}{%
\texttt{research}%
}
\newcommand{\getdiamond}{%
\texttt{ObtainDiamond}%
}

\begin{document}

\maketitle
\newpage 

\begin{abstract}
Reinforcement learning competitions advance the field by providing appropriate scope and support to develop solutions toward a specific problem.
To promote the development of more broadly applicable methods, organizers need to enforce the use of general techniques, the use of sample-efficient methods, and the reproducibility of the results. While beneficial for the research community, these restrictions come at a cost---increased difficulty. If the barrier for entry is too high, many potential participants are demoralized. With this in mind, we hosted the third edition of the MineRL \getdiamond{} competition, MineRL Diamond 2021, with a separate track in which we permitted \textit{any} solution to promote the participation of newcomers. With this track and more extensive tutorials and support, we saw an increased number of submissions. The participants of this easier track were able to obtain a  diamond, and the participants of the harder track progressed the generalizable solutions in the same task.
\end{abstract}

\section{Introduction}
\label{sec:intro}
Reinforcement learning (RL) and deep RL (DRL) have successfully been used to tackle many complex sequential decision-making problems, from board games~\citep{alphazero} to video games~\citep{starcraft2019, berner2019dota} to stratospheric balloon navigation~\citep{bellemare2020autonomous}. Most of these successes required massive amounts of computational resources, mainly because DRL algorithms require millions or billions of environment interactions to train. These algorithms are also sensitive to minor changes in hyperparameters~\citep{henderson2018deep, ibarz2021train} or observations~\citep{cobbe2020leveraging}, leading to higher compute and sample requirements for hyperparameter tuning. Furthermore, solutions often overfit to the domain by exploiting inductive biases, so insights do not transfer to other problems.

Previous DRL competitions mostly focus on achieving the highest possible scores in a particular environment. The winning solutions usually include environment-specific, hardcoded rules and action/reward shaping~\citep{kanervisto2021distilling}. Some competitions even include separate tracks for DRL agents~\citep{kuttler2020nethack, mohanty2020flatland} because they are often weaker than non-learning-based approaches. Another relatively common way to succeed in a DRL competition is to throw massive amounts of compute at the problem~\citep{bou2020pytorchrl}, which not everyone can afford. Such solutions also do not transfer well to real-world settings, where lightweight simulators are not available.

To promote the development of DRL techniques that generalize to new domains and are more sample-efficient, we ran the third iteration of the MineRL Diamond competition as part of the NeurIPS 2021 Competition Track.
The goal of this competition is for agents to obtain a diamond in Minecraft. 
To encourage sample-efficient and generalizable solutions~\citep{houghton2020guaranteeing}, we limited the number of environment interactions agents can be trained on, retrained submissions for final evaluation, and prohibited manual programming of behaviour. 
To encourage more general participation outside of the DRL research community, we introduced a novel track, the \easy{} track, which follows the standard competition format: no limits on hardcoding, training time, or approaches. To increase accessibility, we developed resources, such as Colab notebooks, to help participants get started with training agents. Additionally, we improved our community support through office hours and a workshop.

\section{Competition Overview}
\label{sec:overview}
    To encourage broader participation and move toward the use of the MineRL challenge as a general research benchmark, the \easy{} track acts as a stepping stone towards the main competition. 
    In this track, participants may use any techniques to develop agents, which are evaluated post-training (as is done in standard benchmarks and most other RL competitions).
    We maintained the original competition rules and constraints~\cite{guss2021towards} that foster the development of novel techniques in the \hard{} track.
    We eliminated the imitation-learning-only track from the 2020 competition~\citep{guss2021towards}, since participants tended to leverage a combination of imitation learning and reinforcement learning.
    Focusing on the combined track better reflects the overall goals of the competition.
    
    \subsection{Task}
        The core task, \getdiamond, was to train an agent to obtain a diamond in Minecraft, which requires obtaining resources and crafting tools to progress through the item hierarchy~\citep{guss2019minerl}. Agents start in a random location in a random world generated by Minecraft. The agent has access to a $64 \times 64$ pixel observation from the point-of-view of the player (similar to what a human would see) and its current inventory contents. The agent controls the avatar with actions similar to keyboard buttons (e.g., moving forward or left), selected 20 times per second. The agent receives a reward each time they obtain the next item in the list of required items towards diamond, detailed in \Cref{tab:rewards}. An episode ends once the agent obtains a diamond, the agent dies, or the episode lasts for 18000 steps (15 minutes of in-game time).
        
    \subsection{\hard{} Track}
        Because the \hard{} track retained sample efficiency, reproducibility, and generalizability constraints as in the 2020 competition,
        we provide an outline here and refer to \cite{guss2021towards} for full details.
        The competition structure and rules prohibited manual specification of behaviour. 
        We obfuscated the observations and actions to have no human-readable meaning, which prevented semantic hardcoding. 
        The rules prohibited reward shaping, selecting sub-policies based on environment observations, and manually pruning or selecting actions.
        To enforce these rules, we manually validated the participants' training and evaluation code during both rounds. 
        In case of uncertain rule-breaking, we contacted participants and asked for clarification. Rule-breaking teams were disqualified.
        
        Although in Round 1 participants submitted trained agents along with the training code, in Round 2 the finalists submitted only their training code.
        We trained the submissions on the evaluation server with unseen seeds and environment dynamics to promote generalization.
        To enforce sample efficiency, we limited training to eight million environment interactions on fixed hardware (1 NVIDIA K80 for 4 days).
    
    \subsection{\easy{} Track}
        The \easy{} track had the same challenge without the restrictions of the \hard{} track. Participants were allowed to use any techniques as part of their agent (minus directly modifying the environment), train the agents on any hardware, and only submitted the trained agent. Observations and actions were in the original, human-readable format. However, we still evaluated agents on an unseen set of seeds. To encourage the transition to the \hard{} track, we scheduled the \easy{} track to end a few weeks prior to the end of Round 1 of the \hard{} track. 
    
    \subsection{Resources}
        \begin{wraptable}{r}{0.34\textwidth}
            \small
            \centering
            \caption{Rewards for obtaining items (once per episode).}
            \label{tab:rewards}
            \begin{tabular}{cc}
            \toprule
                \textbf{Item} & \textbf{Reward} \\
            \midrule
                 Log & 1  \\
                 Planks & 2 \\
                 Stick & 4 \\
                 Crafting table & 4 \\
                 Wooden pickaxe & 8 \\
                 Stone & 16 \\
                 Furnace & 32 \\
                 Stone pickaxe & 32 \\
                 Iron ore & 64 \\
                 Iron ingot & 128 \\
                 Iron pickaxe & 256 \\
                 Diamond & 1024 \\
            \bottomrule
            \end{tabular}
        \end{wraptable}

        In addition to the human gameplay dataset~\citep{guss2019minerl}, we provided participants with template code for the submission system for a number of different baseline methods.
        For the \easy{} track, the first baseline used manual behavior specification (a predefined sequence of actions), the second used manual specification in addition to behaviour cloning, and the third used manual specification along with PPO~\citep{schulman2017proximal}.
        These baselines demonstrated how the more flexible rules of \easy{} track can be used advantageously along with learning methods.
        
        For the \hard{} track, we provided a new baseline using behavioural cloning in addition to the baselines from the 2020 competition~\citep{guss2021towards} by Preferred Networks, which were applicable to this competition (four RL methods).
        We also compiled a list of research articles and projects using MineRL in a centralized place as further background material for the participants.\footnote{\url{https://minerl.readthedocs.io/en/latest/notes/useful-links.html}}
        
        To increase the accessibility of the competition, we provided Colab notebook versions of the \easy{} track baselines,\footnote{For example, \url{https://colab.research.google.com/drive/1qfjHCQkukFcR9w1aPvGJyQxOa-Gv7Gt_}} which could be run without any local installations. Together with a video guide on how to make a submission through the browser, the Colab-based baselines enabled people to fully participate in the competition on any device with an internet connection. We also organized a virtual workshop,\footnote{\url{https://minerl.io/workshop}} where participants worked on their submissions or a set of tutorial tasks\footnote{\url{https://github.com/minerllabs/getting-started-tasks}} to get interested people started, regardless of skill.

    \subsection{Evaluation}
        As in the previous years, we hosted the competition with AICrowd. 
        Participants created a git repository to which they uploaded their submission code. The evaluation system cloned the repository, built a Docker image according to the participant's settings, and performed the training and/or evaluation. The Minecraft environment ran in a separate container to prevent the participants from accessing the environment. 
    
        Submissions were evaluated based on the average sum of rewards over 100 episodes.
        Each relevant item for obtaining a diamond provided an exponentially increasing reward, illustrated in \Cref{tab:rewards}.
        This system credited agents that are closer to obtaining a diamond over agents that reliably obtain earlier items to prevent specification gaming~\citep{clark2016faulty}.
        For participants who submitted multiple solutions, the final score was the highest over all of their submissions.

\section{Solutions}
\label{sec:solutions}
    In~\Cref{sec:performance-overview}, we describe the performance of the submissions and how they compare to the performance of submissions to the 2020 edition of the competition. 
    The remaining sections summarize the techniques used by the competitors.

\subsection{\hard{} Track Submissions Overview}
\label{sec:performance-overview}
    
    \begin{figure}
        \centering
        \includegraphics[width=.8\linewidth]{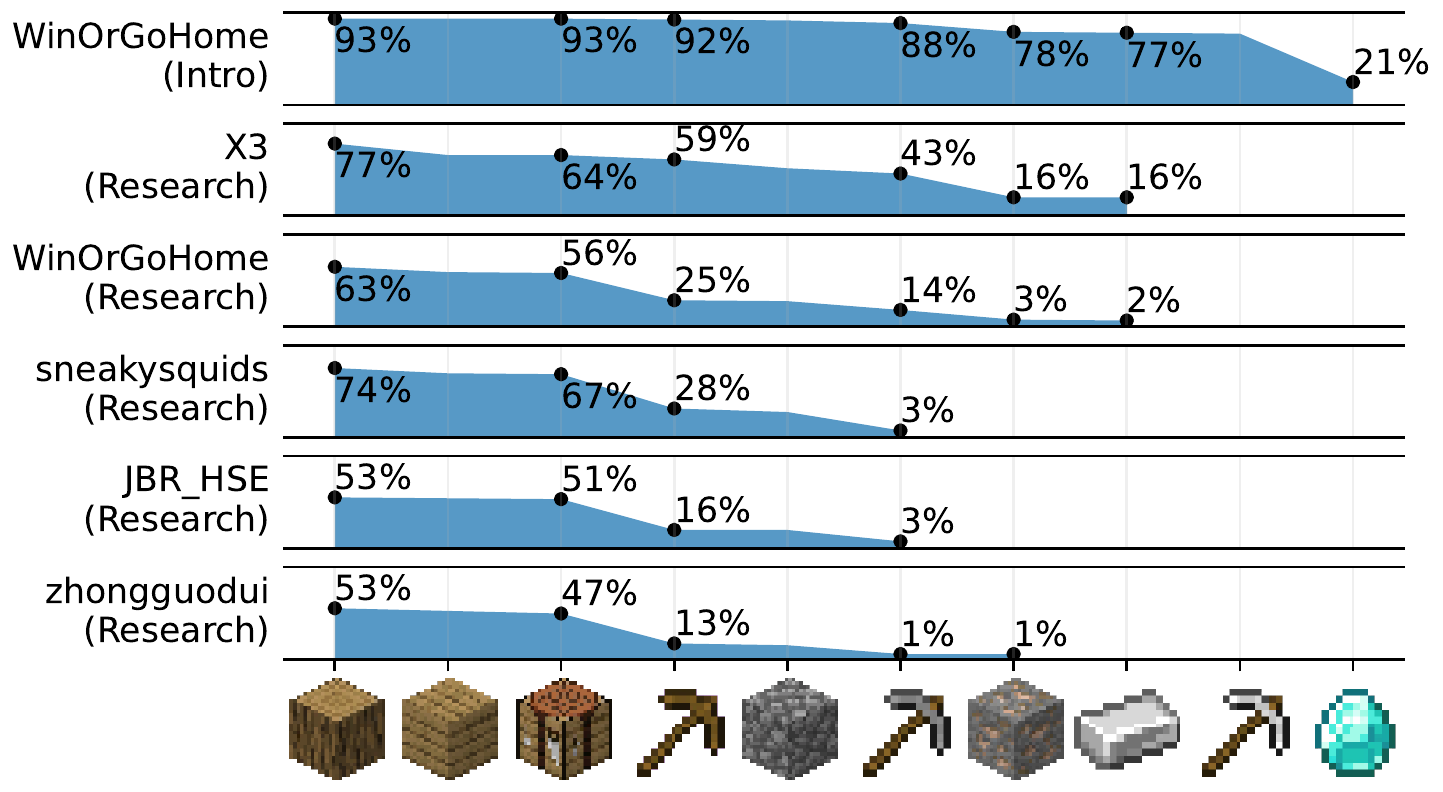}
        \caption{Fraction of episodes agent obtains the item on x-axis, from the Round 2 top submissions for \hard{} track and the top \easy{} submission, measured over 100 episodes.}
        \label{fig:item-distribution}
    \end{figure}

    \begin{table}[]
        \centering
        \caption{Top 10 submissions to both tracks, including submissions to the 2020 edition of the competition for \hard{} track Round 2.}
        \label{tab:leaderboard}
        \begin{tabular}{lrlrlr}
            \toprule
            \multicolumn{2}{c}{\textbf{\easy{}}} & \multicolumn{2}{c}{\textbf{\hard{}, Round 1}} & \multicolumn{2}{c}{\textbf{\hard{}, Round 2}} \\
            \textbf{Name} & \textbf{Score} & \textbf{Name} & \textbf{Score} & \textbf{Name} & \textbf{Score} \\
            \midrule
WinOrGoHome               & 645.55 & X3                            & 68.12      &  X3                        & 76.97          \\
X3                        & 442.38 & PA-P                          & 54.94      &  \textbf{2020 \#1}                  & 39.55          \\
zhongguodui               & 349.97 & SneakySquids                  & 22.54      &  WinOrGoHome               & 22.97          \\
MCAgent                   & 323.82 & MCAgent                       & 17.70       &  SneakySquids              & 14.35          \\
ISYAgent                  & 220.03 & xianyu                        & 16.71      &  \textbf{2020 \#2}                  & 13.29          \\
ced                       & 167.78 & Reforcos\_de\_Minecraft       & 13.36      &  \textbf{2020 \#3}                  & 12.79          \\
xianyu                    & 79.06  & zhongguodui                   & 11.08      &  JBR\_HSE                  & 10.33          \\
PA-P                      & 69.50  & WinOrGoHome                   & 10.55      &  zhongguodui               & 8.84           \\
ttom                      & 45.35  & grimoirez                     & 8.55       &  MikeAI                    & 6.25           \\
orithu                    & 38.42  & okkkkkkkk                     & 6.44       &  \textbf{2020 \#4}                  & 5.16           \\
            \bottomrule
        \end{tabular}
    \end{table}

    \begin{table}[t]
        \centering
        \caption{Competition activity statistics.}
        \label{tab:comparisons}
        \begin{tabular}{cccc}
        \toprule 
        \textbf{Metric} & \textbf{2019} &  \textbf{2020} & \textbf{2021}\\
        \midrule 
         Number of Teams   & 47 & \textbf{95} & 59 \\
         Number of Individuals & \textbf{1115} & 707 & 468 \\ 
         Number of Submissions & 662 & 490 & \textbf{1190} \\
         Non-zero-score Pre-Final-Round Submissions & 38 & 36 & \textbf{55} \\ 
         Non-zero-score \hard{} Track Submissions & \textbf{38} & 36 & 36 \\
         Number of Discord Server Messages & 2464 & 2444 & \textbf{5556} \\
         New Discord Server Users & 445 & 160 & \textbf{955} \\
        \bottomrule 
        \end{tabular}
    \end{table}
    \Cref{tab:comparisons} compares this year's competition activity with previous ones.
    This year, we attracted more active participation: despite a decreased number of teams compared to the 2020 edition and the lowest number of individual competitors, the 2021 edition saw the most submissions. 
    This finding is further supported by the higher number of non-zero-score submissions before the final round for both tracks and a comparable number of non-zero-score submissions in the Round 1 \hard{} track.
    We believe this increased active participation is due to the inclusion of the \easy{} track, which eased the entry to the competition.
    
    In the \easy{} track, 16 teams outperformed the baseline. In the \hard{} track 20 teams outperformed the BC baseline in Round 1.
    In Round 2 of the \hard{} track, nine teams achieved a non-zero score. 
    \Cref{tab:leaderboard} shows the final ranking of the \easy{} and \hard{} tracks, respectively. The overall winner of the \hard{} track outperformed last year's top solution~\citep{mao2021seihai} by a large margin.\footnote{Original \#1 team from the 2020 competition with over 70 score was retroactively disqualified for breaking the rules.}
    
    Solutions in the \easy{} track reached notably higher scores (top three average of 479.3 vs. 48.5 for \hard{} Round 1), due to fewer restrictions on usable techniques. With these less restrictive rules, teams could design useful macro actions and train agents on them (e.g., WinOrGoHome) or hard-code most of the behaviour, with more difficult sub-parts learned by an RL algorithm (e.g., X3).

\subsubsection{\hard{} First Place: Team X3}
\label{sec:x3}
    The first-place team (X3) proposed a sample-efficient hierarchical RL approach equipped with representation learning and imitation learning to deal with perception and exploration~\citep{lin2021juewu}. Specifically, their approach included two levels of hierarchy---the high-level meta-policy learned to control over options to switch between different sub-tasks, while the low-level sub-policies learned to solve each sub-task efficiently. To boost the learning of sub-tasks, they used a combination of techniques including 1) action-aware representation learning which captured underlying relations between action and representation, 2) discriminator-based self-imitation learning for efficient exploration, and 3) ensemble behavior cloning with consistency filtering for policy robustness. These techniques significantly improved the sample efficiency of learning agents and the proposed approach achieved the highest score (76.97) in the research track, which outperformed last year's top solution by a large margin (39.55).

\subsubsection{\hard{} Second Place: Team WinOrGoHome}
\label{sec:winorgohome}
    Team WinOrGoHome based their solution of team NoActionWasted from the previous year~\citep{guss2021towards}. This solution used k-means to cluster the actions, after which a recurrent neural network was trained with behavioural cloning. Team WinOrGoHome improved this solution by following techniques:

    \begin{itemize}
    \item Add the current reward into the state for helping the agent distinguish between different stages.
    \item For helping the agent focus on the key frames, weigh the policy loss by the magnitude of the change in the obfuscated observation vector (this contains information of the player's inventory and item held in hand).
    \item Predict the discounted return on the \texttt{ObtainIronPickaxe} dataset as an auxiliary task, where the target to predict is $\log(1+\text{value})/8$.
    \item Add linear learning rate decay, from $6.25 \cdot 10^{-5}$ to $1.00 \cdot 10^{-5}$.
    \end{itemize}
    
    This improved imitation-only agent ranked 2nd place on the final leaderboard, scoring 22.97, which is a notable improvement over team NoActionWasted's score of 12.79.

\subsubsection{\hard{} Third Place: Team SneakySquids}
\label{sec:sneakysquids}
    Language models~\citep{gpt1, gpt2, gpt3, bert} have demonstrated strong results both in text-based environments~\citep{butlers, calm} and in tasks and environments that have seemingly no ties to natural language processing~\citep{universal, wikipedia}. Team SneakySquids followed this line of work by finetuning a pretrained GPT-S model~\citep{gpt2} with behavioural cloning of expert gameplay in the \texttt{Treechop}, \texttt{ObtainIronPickaxe} and \texttt{ObtainDiamond} environments. During an initial data preprocessing step they removed below average demonstrations and discretized the action space with k-means clustering of expert actions. Then the model was trained to predict the actions taken by expert players conditioned on a sequence of previous observations. An observation consisted of two distinct embeddings: a frame embedding learnt end-to-end by a convolutional neural network~\citep{fixup} and an obfuscated inventory embedding given by the environment.
    
    With this design, the agent had access to a compressed representation of the last ten seconds of gameplay for deciding the action, which enabled smooth navigation and short bursts of crafting actions. Consistently with recent work~\citep{wikipedia}, team SneakySquids observed that a pretrained GPT-2 model learnt faster and achieved higher scores than GPT-2-like architectures initialized from scratch. Even though this approach relied purely on imitation learning and was thus heavily penalized by the scarcity of the training data, their best performing agent achieved an average score of 22.540 on the evaluation platform in Round 1, and 14.35 in the final ranking of Round 2.
    
\subsubsection{\hard{} Fourth Place: Team JBR\_HSE}
\label{sec:jbr}
    Team JBR\_HSE's approach was based on a composition of hierarchical reinforcement learning and imitation learning. They hypothesized that RL performs well with short-horizon planning and high-level actions, while micromanagement is too complex and requires more time to learn than the competition limit provides. Therefore, micromanagement should be learned through imitation learning.

    On a high level, their approach involved two levels, i.e., one high-level policy and many low-level options~\citep{SUTTON1999181}. All trajectories were evenly divided into $N$ chunks by time. Then a separate IL policy was learned as an option for each chunk. After options were trained, they were used in place of direct actions in a Deep Q-Network (DQN)~\citep{mnih2013playing}. When DQN chose an option, the policy responsible for it performed $K$ steps in the environment with low-level actions from the original action space. In all architectures they used standard ResNet~\citep{fixup} and long short-term memory (LSTM)~\citep{hochreiter1997long} (for both levels).
    
    Such an approach appeared to be better than pure imitation learning but did not achieve impressive results (10.33 in Round 2). The main reason for this is that larger intervals of training data resulted in better options but as a result, DQN’s training budget was reduced. This work was supported by computational resources of HPC facilities at HSE University~\citep{kostenetskiy2021hpc}.

\subsection{\easy{} Track Submissions Overview}
\label{sec:intro-performance-overview}
    With the relaxed rules of the \easy{} track, top participants were able to obtain the diamond for the first time in all of the MineRL competitions. While these results are not directly comparable to previous MineRL competitions or \hard{} track submissions, it demonstrates the effectiveness of domain knowledge when it comes to training agents.

\subsubsection{\easy{} First Place: Team WinOrGoHome}
\label{sec:intro-winorgohome}
    Team WinOrGoHome leveraged end-to-end reinforcement learning without any hierarchy training paradigm.
    For the action space, they designed 20 useful discrete actions with some actions that sometimes get masked according to their comprehension of Minecraft, which largely sped up the exploration.
    For feature engineering, they used every fourth observation and modified the image and vector observations. They stacked the most recent four non-skipped frames into a $64\times64\times12$ tensor for the image part. The vector part included the info of masked actions, the current count of items in the inventory together with their cumulative counts through the episode.
    They designed a more elaborate reward system helping the agent explore the hierarchical dependency property of different items.
    Their model follows the structure in IMPALA~\citep{espeholt2018impala} but without long LSTM~\citep{hochreiter1997long}, while PPO~\citep{schulman2017proximal} is used to train the model.
    Their final model achieved an average score of above 560 after consuming 625 million frames (i.e., 2.5 billion frames in total) and ranked 1st for the \easy{} track with the best submission score of 645.55. The agent obtained 21 diamonds out of the overall 100 episodes in this submission.

\subsubsection{\easy{} Second Place: Team X3}
\label{sec:intro-x3}
    For the intro track, team X3 used a hard-coded behavior tree for all stages, except for tree-chopping, where Soft Q Imitation-Learning~\citep{reddy2019sqil} was used instead. Several combo-actions were designed to achieve high-level abstract actions, such as ``attack-until-inventory-changes-with-timeout", ``reset-yaw-pitch" and ``clear-nearby-ore-vein". In order to avoid death and recognize certain blocks, they deployed a simple yet effective hard-coded color-mask detection. The agent gets an iron-pickaxe in 70\% of the runs and gets a diamond in 10\% runs.

\section{Discussion}
\label{sec:discussion}
Here we discuss the impact of the inclusion of the \easy{} track and the progress made in the task over three years of MineRL Diamond competitions.

    \subsection{Impact of the \easy{} Track}
        While the high-scoring solutions of the \easy{} track are not academically as interesting as solutions for \hard{} track (generalizable), the inclusion of \easy{} track along with newcomer-friendly material attracted more newcomers to the field. We saw more activity on our Discord server for the competition along with more new users (see~\Cref{tab:comparisons}).\footnote{Some of the users joined for the MineRL BASALT competition which used the same Discord server, but majority joined for the Diamond competition.} We also received multiple messages thanking us for the more inclusive nature of the competition, noting that getting started was too difficult in the previous years. One participant even reported using the entire inclusivity compute grant and more for this competition.
        
        
        The overall higher score of the \easy{} track solutions compared to the \hard{} track solutions signals that simple, domain-specific engineering can yield more performant agents. For example, crafting in the MineRL environment requires picking a specific action once requirements are satisfied, which can be achieved with a single if-then statement. However, chopping down a tree requires locating a tree and then navigating to a it, which is not easily programmable.
        Instead, learning it with RL has proven more effective. This provides a more complete picture of ``which approaches work" and provides a fairer comparison between symbolic and learning agents. Similar results were observed in the NetHack Challenge~\citep{kuttler2020nethack}, where symbolic agents outperformed learning agents by a considerable margin.\footnote{\url{https://nethackchallenge.com/report.html}}
        
        Given these insights, we highly recommend that competitions aim to include a track similar to the \easy{} track, along with more approachable tutorials, instructions, events and submission instructions, to attract people outside the field of AI. However, this easier track should not replace the original, and more difficult, research track, but should be held alongside it. This way, the easier track helps to attract participants and drive the competition by including different parties. It will also create a sense of progression which helps with the morale of the teams in the harder track. The harder track should be encouraged by providing larger prizes or by ending the easier track sooner to allow people to switch tracks. We also advise preparing answers for newcomers to programming, especially if you organize a competition around a video game. We received messages from many users who were familiar with Minecraft but did not have experience with programming but were eager to try out MineRL. We directed these people to the beginner programming tutorials and provided step-by-step code examples for MineRL. We recommend other competitions to do the same, time permitting, to avoid distinguishing the curiosity of interested minds.
    
    \subsection{Brief Analysis of the Results}
        For three years, hundreds of participants have tuned their agents to find the diamond in Minecraft, each year with their own setting. In 2019, the competition outline was designed and focused on the sample efficiency and reproducibility of the training results~\citep{milani2020retrospective}. In 2020, this was further encouraged with the obfuscated observations and actions~\citep{guss2021towards}. In 2021, we noted that the restrictions limit active participants to a select group (highly skilled people in RL) and added a less-constrained track to ramp up newcomers. Only in this easier track and after two competitions, did we finally see agents obtaining diamonds. How come?
        
        \textbf{Rarity of important events.} For example, to obtain a wooden pickaxe (\Cref{fig:item-distribution}, fourth item), the player must craft a crafting table (third item), place it down, look towards it and execute ``craft wooden pickaxe" action. If the player has the necessary items in the inventory, they will be replaced with a wooden pickaxe. This only needs to be done once per episode to progress further. Despite getting an immediate reward for crafting the item, this is difficult to learn for the RL agent to due its rarity (once per episode) and the dependencies it requires. However, manually programming this behaviour is simple. This is reflected in \Cref{fig:item-distribution}, where the \hard{} track agents see a significant drop at obtaining wooden pickaxe, while \easy{} track agent does not experience this. The research-oriented submissions have, in the three years, largely utilized some form of hierarchical learning to combat this.
        
        \textbf{Strict limits.} With the compute limit of eight million environment steps and an episode time limit of 18,000 steps, participants had 450 full episodes of data to train on.  Sample efficiency is desirable, but this might be too restrictive to obtain high performance, especially with the above limitations. Team WinOrGoHome obtained diamond multiple times by training for longer with crafted actions to avoid the rarity issue in the \easy{} track.
        
        \textbf{MineRL's slow sampling.} Because of the relatively slow sampling rate of MineRL (max. 300 steps-per-second per environment which uses multiple cores), extensive hyperparameter tuning and experimentation requires a large amount of compute. By creating a Minecraft-like game from the ground up, \citet{hafner2021benchmarking} was able to deeper explore the effectiveness of different algorithms. 
        This highlights how important hyperparameter tuning is for good performance, but at the same time partially breaks the desiderata for sample-efficient algorithms.

\section{Related Work}
\label{sec:relatedwork}

The year 2021 saw a number of Minecraft competitions that were organized alongside this MineRL Diamond competition. A sister competition, MineRL BASALT~\citep{shah2021minerl}, made use of the same MineRL library to express desired agent goals with human demonstrations instead of the rewards, and then used human evaluators to rate the agents on how well they completed the tasks. The IGLU competition~\citep{kiseleva2021neurips} used the building mechanics of Minecraft for two challenges for the participants: build an agent that provides descriptions on what to build, and then build an agent which builds a structure based on the instructions. Prior to 2021, MarL\"O competition~\citep{perez2019multi} challenged participants to solve multiple tasks, some unseen, in a multi-agent setting in Minecraft. The Generative Design in Minecraft competition~\citep{salge2020ai} challenges participants to generate settlements that fit into the surrounding environment.

The Procgen competition~\citep{mohanty2021measuring}, the Nethack challenge~\citep{kuttler2020nethack} and the Unity Obstacle Tower challenge~\citep{juliani2019obstacle} all made use of the procedural generation of levels as well. However, only the Procgen competition also limited the number of environment samples for training and retrained the agents on the evaluation server.

As for providing an easier track, the ViZDoom competitions in 2016 and 2017 included a ``limited deathmatch" and ``full deathmatch", where the former took place in known maps and the latter in unknown maps~\citep{wydmuch2018vizdoom}. However, participants were not limited by a number of samples or by strict rules, and the top teams often reported using a large amount of compute.

\section{Conclusion}
\label{sec:conclusion}
We ran the 2021 MineRL Diamond competition to promote the development of general, sample-efficient DRL and imitation learning algorithms. In this work, we described the competition, highlighted changes from the previous competitions, and summarized the performance of the submissions, and contrasted them with the performance from last year. 
The core changes this year were the inclusion of an easier track and more tutorials to welcome newcomers. We saw an increased number of submissions and active participants, and the top team outperformed the previous best result by a large margin.

These results demonstrate that hosting an easier track of a more challenging competition is beneficial, and we recommend other competitions take this approach. In the future, the core of the MineRL challenge could be redesigned to reap the most benefit of a public competition, and the results over three years could be summarized and analysed in detail.

\acks{
We thank the other organizers for helping to run the competition: William H. Guss, Brandon Houghton, Byron Galbraith, Noboru Sean Kuno, Rohin Shah, Steven H. Wang, and Cody Wild.
We also thank the advisory committee.
We thank Microsoft and AI Journal for providing financial support. 
We are grateful to AICrowd members Sharada Mohanty and Shivam Khandelwal for setting up the backbone of the submission, evaluation and ranking system. Last but not the least, we would like to thank all the participants on the MineRL Discord server who helped others by answering questions, sharing thoughts, and reporting their progression.
}

\bibliography{ref}

\end{document}